\documentclass[letterpaper, 10 pt, conference]{ieeeconf}  

\IEEEoverridecommandlockouts

\usepackage{amsmath,amsfonts}
\usepackage{graphicx}
\usepackage{bm}
\usepackage{soul,color}
\usepackage{multirow}

\title{\LARGE \bf
CorrA: Leveraging Large Language Models for Dynamic Obstacle Avoidance of Autonomous Vehicles
}

\author{Shanting Wang$^{1}$, Panagiotis Typaldos$^{2}$ and Andreas A. Malikopoulos$^{3}$
\thanks{*This research was supported in part by NSF under Grants CNS-2401007, CMMI-2219761, IIS-2415478, and in part by MathWorks.}
\thanks{$^{1}$Shanting Wang is with the System Engineering Program,
        Cornell University, Ithaca, New York State, United States 
        {\tt\small sw997@cornell.edu}}%
\thanks{$^{2}$Panagiotis Typaldos is with the School of Civil and Environmental Engineering,
        Cornell University, Ithaca, New York State, United States
        {\tt\small \{pt432\}@cornell.edu}}%
\thanks{$^{3}$Andreas A. Malikopoulos is with the School of Civil and Environmental Engineering and the System Engineering Program,
        Cornell University, Ithaca, New York State, United States
        {\tt\small \{amaliko\}@cornell.edu}}%
}
\begin{document}

\maketitle
\thispagestyle{empty}
\pagestyle{empty}

\begin{abstract}
In this paper, we present \textit{Corridor-Agent (CorrA)}, a framework that integrates large language models (LLMs) with model predictive control (MPC) to address the challenges of dynamic obstacle avoidance in autonomous vehicles. Our approach leverages LLM reasoning ability to generate appropriate parameters for sigmoid-based boundary functions that define safe corridors around obstacles, effectively reducing the state-space of the controlled vehicle. 
The proposed framework adjusts these boundaries dynamically based on real-time vehicle data that guarantees collision-free trajectories while also ensuring both computational efficiency and trajectory optimality. The problem is formulated as an optimal control problem and solved with differential dynamic programming (DDP) for constrained optimization, and the proposed approach is embedded within an MPC framework. Extensive simulation and real-world experiments demonstrate that the proposed framework achieves superior performance in maintaining safety and efficiency in complex, dynamic environments compared to a baseline MPC approach.
\end{abstract}

\section{Introduction}
The rapid development of advanced sensing, computation, and artificial intelligence technologies has made autonomous vehicles (AVs) more realistic and made related studies unprecedented. However, the complexity, dynamics, and unpredictability of real-world environments have impeded the deployment of AV applications. Until AVs dominate the transportation market, we face the challenge of mixed autonomy systems where AVs and human-driven vehicles (HDVs) must coexist safely. The unpredictable driving behavior of HDVs brings uncertainty, which requires AVs to generate safe, efficient, and dynamically feasible trajectories in path-planning tasks. Recently, a variety of methods have been proposed by researchers to address this challenge, including Hamiltonian analysis \cite{malikopoulos2021optimal, malikopoulos2018decentralized,mahbub2020sae-1}, reinforcement learning (RL) \cite{kiran2021deep,aradi2020survey, nakka2022multi}, and model predictive control (MPC) \cite{Le2023ACC,le2024controller,guanetti2018control, typaldos2022optimization} approaches. Although Hamiltonian analysis approaches yield analytical solutions, they struggle with complex, real-world constraints. RL approaches learn optimal policies through interactions with the environment and have shown great capability in managing the complexities and uncertainties of dynamic systems. However, RL methods often face limitations due to insufficient training data during the exploration phase. Standard MPC approaches, despite their predictive capabilities, often feature high computational demands and may struggle with complex, large-scale optimization problems.

One critical question that still remains unanswered is ``How can we ensure safe trajectory planning under uncertain traffic conditions?" To address this question, we propose \textit{CorrA}, a hybrid framework that combines MPC with large language models (LLMs) to define adaptive sigmoid-based boundary functions. This integration enables autonomous vehicles to navigate complex mixed-autonomy traffic environments with enhanced computational efficiency and safety guarantees. Unlike prior methods utilizing potential functions, our approach considers the defined boundaries as hard constraints for collision avoidance. Several studies have examined the formation of sigmoid functions for collision avoidance but are limited to simple environments like two-lane roads \cite{lu2020hybrid, ammour2021collision}. We uniquely leverage LLMs' reasoning capabilities to determine appropriate parameters for sigmoid-based boundaries around obstacles dynamically. We then formulate an optimization problem, which is solved using differential dynamic programming (DDP) \cite{murray1979constrained}, considering these boundaries as hard constraints, and the whole procedure is embedded within an MPC framework. To evaluate our method, we consider numerous realistic scenarios on multi-lane road segments and compare them against a baseline MPC approach. The results indicate that CorrA outperforms the baseline across multiple areas, such as computational and traffic-related efficiency.

The main contributions of this paper are:
   (1) The development of a boundary formulation that replaces traditional potential functions with sigmoid-based boundaries for obstacle avoidance in complex environments. By creating explicit boundaries, the state space of the controlled vehicle is reduced, which could lead to faster computation times and more reliable performance compared to potential field methods.
    (2) The development of a hybrid framework combining real-time efficiency and safety guarantees of MPC with the adaptability of LLMs in dynamic environments.
    We provide simulation results and scaled experiments that demonstrate that the proposed hybrid approach outperforms a baseline MPC approach. 

\begin{figure*}[h]
    \centering
      \includegraphics[width=0.95\textwidth]{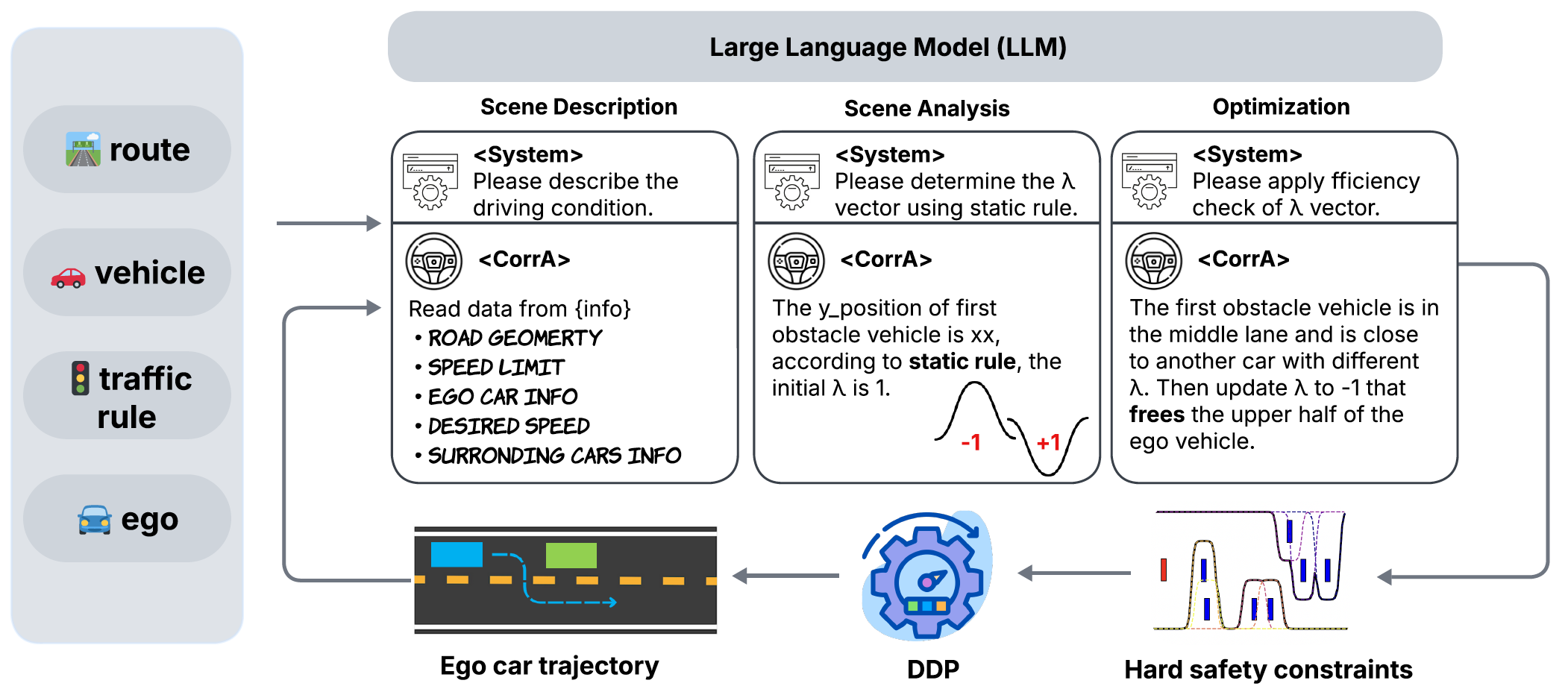}
    \caption{Pipeline of the \textit{CorrA}. We input the vehicles' dataset into LLM to get the driving condition. The LLM will define the initial $\lambda$ values for the obstacle vehicles using static rules. Then, LLM will perform a rule-based travel time efficiency check to update the $\lambda$ values, aiming to maximize the ego car's free space. Then, the safety hard constraints for the ego car are decided based on the $\lambda$ values. Subsequently, we solve the optimization problem embedded within an MPC framework to obtain the vehicle's optimal trajectories. Ultimately, we update the vehicle information for the next round of path planning.}
    \label{fig:pipeline}
\end{figure*}

\section{Related Work}

\subsection{Large Language Model}
The integration of LLMs into autonomous driving systems has emerged as a significant research direction. Approaches are broadly categorized into three methodological frameworks: prompt engineering, fine-tuning, and multi-modal. The authors in \cite{wen2023dilu} introduced a framework that incorporates memory modules to facilitate experience-based reasoning. A system enabling conversational interaction was developed in \cite{cui2024drive} that allows for the reception of human verbal commands, subsequently processing these inputs to dynamically influence vehicle behavior. The authors in \cite{sha2023languagempc} proposed a method that converts high-level driving decisions into parameter representations that guide a low‑level MPC system. While these approaches demonstrate the potential of LLMs in driving contexts, they primarily utilize LLMs as interpretation or translation layers between human commands and vehicle control systems rather than leveraging LLMs' reasoning capabilities to formulate optimization constraints.

TrafficGPT developed in \cite{zhang2024trafficgpt} bridges the gap between natural language understanding and domain-specific traffic analysis. The authors in \cite{chen2024driving} implemented object‑level vector representations within an LLM framework to enhance the explainability of driving decisions. A policy adaptation mechanism was proposed in \cite{li2024driving} that adapts traffic rules to new environments. Mao et al. \cite{mao2023gpt} reformulated motion planning as a language modeling problem. The authors in \cite{xu2024drivegpt4} focused on the processing of multi-modal input data and generating low-level control signals for end-to-end autonomous driving applications. Beyond this, Tian et al. \cite{tian2024drivevlm} extended this framework by integrating the vision-language model to improve scene-understanding capabilities. Choudhary et al. \cite{choudhary2024talk2bev} explored this direction further by introducing an interface specifically designed for bird’s-eye view maps. While these approaches offer promising directions, they may lack safety guarantees that are critical in mixed-autonomy traffic environments. Furthermore, unlike previous approaches that often require extensive training datasets or fine-tuning, our method utilizes pre-trained LLMs as reasoning engines that can be deployed immediately in diverse traffic scenarios without additional training.

\subsection{Differential Dynamic Programming}

Differential Dynamic Programming (DDP), introduced in \cite{mayne1970} and later extended by \cite{murray1979constrained, murray1984differential}, is an iterative algorithm that progressively improves trajectories until converging to an optimal control solution. Each iteration computes a quadratic-linear approximation of the recursive Bellman equation around the current trajectory.  

DDP has been applied in several areas, including robotics, unmanned aerial vehicles (UAV), connected and automated vehicles (CAVs), multi-agent systems, and recently, urban air mobility. Specifically, the authors in \cite{tassa2014control} address the challenge of enforcing actuator and torque limits while retaining the computational efficiency of DDP. Their main application is real-time whole-body humanoid control: they demonstrate the algorithm on the HRP-2 robot, showcasing dynamically feasible motions like balancing and reaching while respecting strict control bounds. The authors in \cite{xie2017differential} extended DDP to handle arbitrary nonlinear constraints on both states and controls. The approach was demonstrated using UAVs that maneuver around obstacles, verifying that the DDP can maintain feasibility even with dynamic constraints. Two control frameworks of decentralized multi-decision-makers \cite{Malikopoulos2021} that combine DDP with the ADMM algorithm to address large-scale problems involving hundreds of agents (e.g., CAVs or UAVs) were introduced in \cite{saravanos2023distributed}. They demonstrate impressive results in simulation (e.g., coordinating 1024 cars) and validate the approach with hardware experiments on a real multi-robot platform. In a different context, the authors in \cite{typaldos2023modified} focused on the green light optimal speed advisory (GLOSA) problem, considering adaptive signals, and achieved real-time numerical solutions for vehicles approaching traffic signals using discrete differential dynamic programming and DDP. However, the majority of the works have focused on static environments, while the current approach addresses a highly dynamic environment where surrounding vehicles continuously change their speed and position.

\section{Problem Formulation}\label{sec: problem}
The current work addresses the problem of autonomous driving in dynamic, mixed-autonomy traffic environments. Consider an autonomous vehicle (AV) driving in a multi-lane roadway, where surrounding traffic consists of both autonomous and human-driven vehicles (HDVs). The goal is for the AV to navigate safely and efficiently while adapting to the dynamically changing surrounding traffic.

\subsection{Overview}
The key idea of CorrA is to define dynamic safety boundaries that create a collision-free region for the controlled AV by leveraging the reasoning capabilities of pre-trained LLMs.

\begin{figure}[t]
      \centering
      \includegraphics[width=\linewidth]{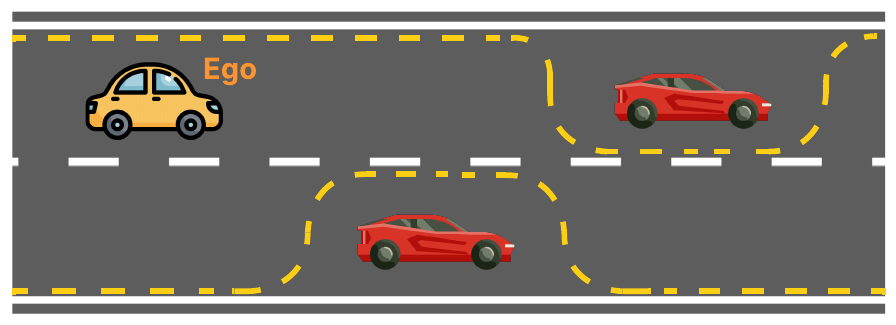}
      \caption{Example of ``safe" corridor's lower and upper boundaries (dashed yellow lines) follows the sigmoid safety constraints.}
      \label{corridor_sigmoid}
   \end{figure}

Our three-phase pipeline proceeds as follows (see Fig. \ref{fig:pipeline}): 
\begin{enumerate}
    \item We first collect real-time, updated data for each vehicle and corresponding infrastructure geometry information.
    \item We input driving conditions into an LLM (e.g., o3-mini) to generate the $\lambda$ values for each obstacle vehicle with LLM's reasoning interpretations. These $\lambda$ values are subsequently used to formulate the hard safety constraints applicable to the ego vehicle.
    \item The LLM output is given as input to the trajectory optimization controller, which solves an optimal control problem (OCP) over a specific time horizon. The OCP is embedded within an MPC framework to ensure real-time, robust trajectory planning that can adapt to dynamic constraints and uncertainties.
\end{enumerate}

\subsection{Vehicle Dynamics and Constraints}

The discrete-time vehicle dynamics consider both lateral and longitudinal directions and are described by a double integrator model as
\begin{align}
    x_{k+1} &= x_k + v_{x,k} \cdot T + \tfrac{1}{2}\,u_{x,k}\cdot T^2, \label{eq: state-x} \\
    y_{k+1} &= y_k + v_{y,k}\cdot T + \tfrac{1}{2}\,u_{y,k}\cdot T^2, \label{eq: state-y} \\
    v_{x,k+1} &= v_{x,k} + u_{x,k}\cdot T, \label{eq: state-vx} \\
    v_{y,k+1} &= v_{y,k} + u_{y,k}\cdot T, \label{eq: state-vy}
\end{align}
where the state variables $x_k, y_k, v_{x,k}, v_{y,k}$ represent the longitudinal and lateral positions and speeds at discrete time-step $k$, respectively, while $u_{x,k}, u_{y,k}$ are the control variables reflecting on the longitudinal and lateral accelerations. Note that the control variables are kept constant for the duration $T$ of each time-step $k$.

The control variables $\bm{u}_k = [u_{x,k}, u_{y,k}]^T$ are bounded according to the specifications and restrictions of the vehicle as follows:
\begin{align}
    u_{x,L}(\bm{x}_k) \leq u_{x,k} \leq u_{x,U}, \label{eq: bounds_x} \\
    u_{y,L}(\bm{x}_k) \leq u_{y,k} \leq u_{y,U}(\bm{x}_k), \label{eq: bounds_y}
\end{align}
where $\bm{x}_k = [x_k, y_k, v_{x,k}, v_{y,k}]^T$ is the states vector. 

Specifically, in the longitudinal direction, the upper bound in \eqref{eq: bounds_x} is constant, i.e., $u_{x,U} = u_x^{\max}$, and reflects the acceleration capabilities of the vehicle. On the other hand, the lower bound is designed appropriately as a state-dependent bound
\begin{equation} \label{eq: bound_y_low}
    u_{x,L}(\bm{x}_k) = \max\left\{-\frac{1}{T}v_{x,k}, u_x^{\min}\right\}.
\end{equation}

The last equation guarantees that the vehicle does not reach negative longitudinal speed values, and its lower value is greater than or equal to a constant minimum value $u_x^{\min}$.

The lateral acceleration bounds are designed appropriately to fulfill two key objectives: (i) to ensure that the controlled vehicle remains within the road boundaries, and (ii) to form a safe region around the obstacles by excluding them from the feasible area of the controlled vehicle (see Fig. \ref{corridor_sigmoid}).

Thus, the derived upper and lower bounds in \eqref{eq: bounds_y} are given as follows
\begin{equation}
\begin{aligned}
    u_{y,U}(\bm{x}_k) = \dfrac{2( (\tilde{y}_l(\bm{x}_k) - y_k) - v_{y,k} \cdot T )}{T^2},
\end{aligned}
\end{equation}
\begin{equation}
\begin{aligned}
    u_{y,L}(\bm{x}_k) = \dfrac{2( (\tilde{y}_r(\bm{x}_k) - y_k) - v_{y,k} \cdot T )}{T^2},
\end{aligned}
\end{equation}
where $\tilde{y}_r(\bm{x}_k)$ and $\tilde{y}_l(\bm{x}_k)$ are the lower and upper lateral positions. The $\tilde{y}_r(\bm{x}_k)$ and $\tilde{y}_l(\bm{x}_k)$ are designed as the summation of two sigmoid functions for each obstacle vehicle as follows
\begin{equation} \label{eq: sigmoid_l}
\begin{aligned} 
    \tilde{y}_l(\bm{x}_k) = \begin{cases}
    r_w - \dfrac{r_w - (y_{o,i} + l_w)}{1 + e^{-slp(x(k) - swp1)}} \\
                   \qquad + \dfrac{r_w - (y_{o,i} + l_w)}{1 + e^{-slp(x(k) - swp2)}} & \text{if \quad} \lambda = 1\\
                    r_w - \dfrac{l_2}{2} & \text{if} \quad \lambda = -1
    \end{cases}
\end{aligned}
\end{equation}
\begin{equation} \label{eq: sigmoid_r}
\begin{aligned}
    \tilde{y}_r(\bm{x}_k) = \begin{cases}
    \dfrac{l_w}{2} & \text{if} \quad \lambda = 1 \\
        \dfrac{r_w - y_{o,i} + (l_w/2)}{1 + e^{-slp*(x(k) - swp1)}} \\
                   \qquad \quad + \dfrac{r_w - y_{o,i} + (l_w/2)}{1 + e^{-slp(x(k) - swp2)}}  & \text{if} \quad \lambda = -1 
    \end{cases}
\end{aligned}
\end{equation}
where $r_w$ and $l_w$ are the road and lane widths, respectively, while $y_{o_i}$ indicates the lateral position of the obstacle. The boundary functions are either equal to the corresponding road boundaries; or are composed of two sigmoids centered at the positions $swp1$ and $swp2$, representing the front and rear sides of the obstacle vehicle, and the parameter $slp$ controls the steepness. The parameter $\lambda$ decides the orientation of these boundaries and is explained in details in the following section. This formulation enables the dynamic adaptation of safety corridors based on real-time traffic conditions, road geometry, and predicted vehicle trajectories, substantially reducing the feasible state-space for optimization while ensuring collision-free navigation.

\begin{figure}[t]
    \centering
    \includegraphics[width=0.95\linewidth]{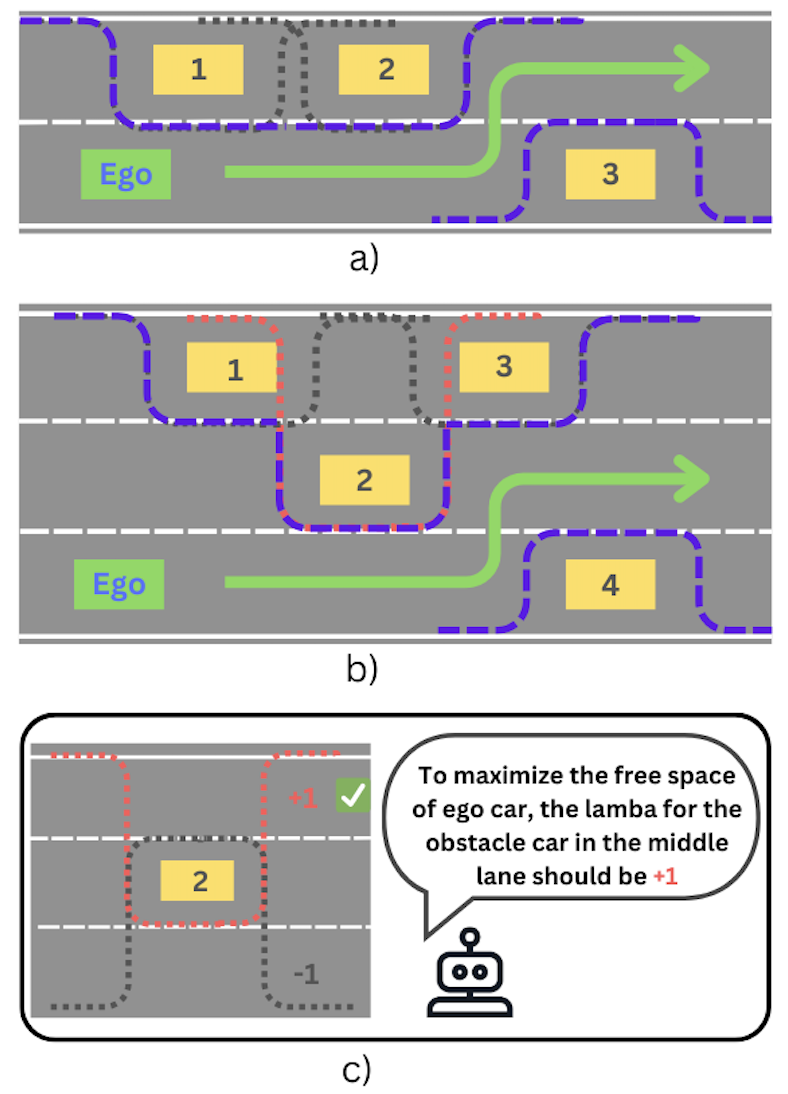}
    \caption{Example of a) two-lane scenario, b) three-lane scenario, and c) the LLMs reasoning process of efficiency check.}
    \label{2/3lane}
\end{figure}

\subsection{Obtain $\lambda$ via Large Language Models}

Reasoning ability is required to decide the appropriate sigmoid function for each obstacle vehicle in a complex scenario. In Fig. \ref{2/3lane} (a), there are three obstacle vehicles. We want to find a space where the ego vehicle can move freely without collision. Vehicles 1 and 2 are close to the upper road boundary, so the $\lambda$ should be set to +1, while vehicle 3 is close to the lower road boundary, meaning the $\lambda$ should be set to -1.

What if there are three lanes and the obstacle vehicle is in the middle lane? The $\lambda$ should be decided depending on the specific scenario. In Fig. \ref{2/3lane} (b), the $\lambda$ of middle vehicle 2 can be either -1 or +1. It is easy for humans to recognize that the $\lambda$ should be positive 1 so that the ego vehicle has more free space. However, how can the vehicle itself decide $\lambda$ value in a dynamic environment?

To address this problem, we leverage the reasoning ability of LLMs to define the $\lambda$ values of obstacle vehicles, as shown in Fig. \ref{2/3lane} (c). The process is illustrated in Fig. \ref{fig:pipeline}: We input the raw data file into the LLM. This file includes current states of both obstacle and ego vehicles. The LLM extracts the key information to describe the driving condition, including:
longitudinal and lateral positions, speeds and accelerations of all vehicles, road geometry (ex. number of lanes and the width of lane), speed limit of the road, and desired speed of the ego vehicle. Then, the LLM determines $\lambda$ values through static rules and efficiency checks. \textbf{Static Rules}: If the obstacle lateral position is greater than or equal to half of the total width of the road, then the $\lambda$ is positive (+1); otherwise, the $\lambda$ is negative (-1). \textbf{Efficiency Check}: If there are more than two lanes, the LLM $\lambda$ perform an efficiency check to decide the lambda values of vehicles. This check ensures that the free space available to the ego vehicle is maximized. The final output of the LLM is a vector of $\lambda$ values with size $1 \times n$, where $n$ represents the number of surrounding obstacle vehicles.

The $\lambda$ values directly influence the shape and behavior of the sigmoid functions defined in ~\eqref{eq: sigmoid_l} and \eqref{eq: sigmoid_r}. A positive $\lambda$ (+1) creates a sigmoid barrier that prevents the ego vehicle from passing above the obstacle, while a negative $\lambda$ (-1) creates a barrier preventing passage below the obstacle. By choosing the appropriate $\lambda$ for each obstacle, we create a feasible space for the ego vehicle. This approach enables dynamic, context-aware decision-making for autonomous vehicles navigating complex multi-lane scenarios with multiple obstacles.

\subsection{Objective Function}
The optimization criterion to be minimized over a time horizon of $K$ steps in the future is described as follows
\begin{equation} \label{eq: cost}
\begin{aligned}
    J &= \sum_{k=0}^{K-1} \Bigg\{ \frac{1}{2}w_1  u_{x,k}^2 + \frac{1}{2}w_2 u_{y,k}^2 + \\
        & \frac{1}{2}w_3  \left(v_{x,k} - v_{d,x}\right)^2 + \frac{1}{2}w_4  \left(v_{y,k} - v_{d,y}\right)^2 \Bigg\},
\end{aligned}
\end{equation}
where $w_1,\dots,w_4$ are the corresponding nonnegative weighting factors of each penalty term. The first two quadratic penalty terms correspond to passenger comfort and fuel consumption by penalizing excessive longitudinal and lateral acceleration values \cite{typaldos2020minimization}. The next two terms reflect the vehicle's advancing goals. Specifically, these terms penalize speed deviations from pre-specified desired values for both longitudinal and lateral speeds. The desired longitudinal speed $v_{d,x}$ is positive and depends on the preference of the passenger and the vehicle type, while the value of the desired lateral speed $v_{d,y}$ is specified according to the different objectives of the vehicle.

\section{Differential Dynamic Programming} \label{sec:ddp-general}
Having defined the sigmoid boundary constraints and the OCP in Section \ref{sec: problem}, our approach employs the DDP algorithm to solve the constrained optimization problem. DDP is a second-order optimization method for solving nonlinear optimal control problems without requiring discretization of the state and control spaces. In this work, we employ an extension of DDP that accounts for inequality constraints.

Consider a discrete-time optimal control problem with states $\bm{x}$ and controls $\bm{u}$ with the recursive Bellman equation 
\begin{align}
    V_k(\bm{x}_k)&=\min_{\bm{u}_k} Q_k(\bm{x}_k, \bm{u}_k), \\
    Q_k(\bm{x}_k, \bm{u}_k)&=L(\bm{x}_k,\bm{u}_k) + V_{k+1}(\bm{f}(\bm{x}_k,\bm{u}_k)) \label{eq: bellman},  
\end{align}
where $V_k$ is the optimal cost function for $k=K-1,\dots,0$, $L$ is the objective function of the OCP \eqref{eq: cost}, $\bm{f}$ is the right-hand side of the state equations \eqref{eq: state-x}-\eqref{eq: state-vy}, and the minimization must be carried out for all feasible controls $\bm{u}_k$ that satisfy any inequality constraints present. The DDP procedure performs, at each time step of each iteration, a quadratic approximation of the term to be minimized in the recursive Bellman equation \eqref{eq: bellman}. The quadratic approximation is taken around nominal trajectories ($\bm{\bar{x}}_k, \bm{\bar{u}}_k$), which are the initial trajectories of each iteration. Define $\delta \bm{x}_k = \bm{x}_k- \bm{\bar{x}}_k$ and $\delta \bm{u}_k = \bm{u}_k - \bm{\bar{u}}_k$. The aim is to find the optimal control law for $\delta \bm{{u}}_k$, which minimizes the quadratic approximation, subject to some inequality constraints. The procedure of the DDP algorithm for each iteration is described in brief as follows (see \cite{yakowitz1986stagewise, murray1979constrained} for more details). 

\subsection{Backward Pass}
During the \textit{backward} pass the quadratic approximation of $Q_k(\bm{x}_k,\bm{u}_k)$ around a nominal point ($\bm{\bar{x}}_k,\bm{\bar{u}}_k$), is given by
\begin{equation}
\begin{aligned}
    Q_k(\delta&\bm{x}_k,\delta\bm{u}_k) \approx\ 
Q^T_{\bm{x},k} \delta\bm{x}_k + Q^T_{\bm{u},k} \delta\bm{u}_k
+ \dfrac{1}{2} \delta\bm{x}_k^T Q_{\bm{x}\bm{x},k} \delta\bm{x}_k \\
&+ \delta\bm{x}_k Q_{\bm{x}\bm{u},k} \delta\bm{u}_k + \dfrac{1}{2} \delta\bm{u}^T_k Q_{\bm{u}\bm{u},k} \delta\bm{u}_k,
\end{aligned}
\end{equation}
where $Q_{\cdot,k}$ are the coefficient matrices, derived as follows
\begin{align}
Q_{x,k} &= L_{\bm{x}} + \bm{f}_x^T V_{x,k+1}, \\
Q_{u,k} &= L_{\bm{u}} + \bm{f}_u^T V_{x,k+1},\\
Q_{xx,k} &= L_{xx} + \bm{f}_x^T V_{xx,k+1} \bm{f}_x + V_{x,k+1} \bm{f}_{xx},\\
Q_{ux,k} &= L_{ux} + \bm{f}_u^T V_{xx,k+1} \bm{f}_x + V_{x,k+1} \bm{f}_{ux},\\
Q_{uu,k} &= L_{uu} + \bm{f}_u^T V_{xx,k+1} \bm{f}_u + V_{x,k+1} \bm{f}_{uu},
\end{align}
where $V_{xx}$ and $V_{x}$ are the Hessian matrix and the gradient vector, respectively. The derivatives of the cost and state equations are computed at the nominal point $(\bm{\bar{x}}_k, \bm{\bar{u}}_k)$.

The unconstrained optimal control perturbation is then given by
\begin{equation}
\delta\bm{u}^*_k = \bm{k}_k + \bm{K}_k\delta\bm{x}_k,
\end{equation}
where
\begin{equation}
\begin{aligned}
    \bm{k}_k &= -Q_{uu,k}^{-1}Q_{u,k}, \\ 
    \bm{K}_k &= -Q_{uu,k}^{-1}Q_{ux,k}.
\end{aligned}
\end{equation}
This minimization outcome is substituted in the quadratic approximation $Q$ to obtain, from the Bellman equation, the approximate optimal value function
\begin{equation}
V_k\left(\bm{x}(k)\right) = Q_k\left(\bm{x}_k,\bm{k}_k+\bm{K}_k\delta\bm{x}_k\right).	
\end{equation}

\subsection{Forward Pass}
The nominal trajectories are updated as
\begin{align}
&\bm{u}_k = \bar{\bm{u}}_k + \varepsilon\Bigl[\bm{k}_k+\bm{K}_k\,\delta\bm{x}(k)\Bigr], \label{eq:forward1}\\
&\bm{x}_{k+1} = \bm{f}\Bigl(\bm{x}_k,\bm{u}_k\Bigr), \label{eq:forward2}\\
&\bm{x}_0: \text{given initial states}, \label{eq:forward3}
\end{align}
with a step-size \(0<\epsilon\le1\) chosen via line search. Convergence is declared when
\begin{equation} \label{eq:convergence}
\left(\sum_{k=0}^{K-1}\|\bm{u}(k)-\bar{\bm{u}}(k)\|^2\right)^{1/2} < \varepsilon_1,
\end{equation}
with \(\varepsilon_1>0\) a small threshold.

\subsection{Constrained DDP}
The standard DDP method \cite{mayne1970} does not account for constraints; thus, to meet the needs of the problem presented in this work, extensions of DDP using nonlinear constraints are utilized \cite{yakowitz1986stagewise, xie2017differential}.

The control constraints are given by
\[
\bm{c}\bigl(\bm{x}_k,\bm{u}_k\bigr) \le \bm{0}.
\]
Let \(\widetilde{\bm{c}}(\bm{x},\bm{u})\) be the linearization of \(\bm{c}(\bm{x},\bm{u})\) about \((\bar{\bm{x}}_k,\bar{\bm{u}}_k)\). The quadratic programming subproblem in the backward pass is
\begin{equation}
\begin{aligned}
\min_{\delta\bm{u}_k} \quad & Q\Bigl(\bar{\bm{x}}_k,\bar{\bm{u}}_k+\delta\bm{u}_k\Bigr)\\[1mm]
\text{s.t.} \quad & \widetilde{\bm{c}}\Bigl(\bar{\bm{x}}_k,\bar{\bm{u}}_k+\delta\bm{u}_k\Bigr) \leq \bm{0}.
\end{aligned}
\end{equation}
Let \(S\) denote the active set of constraints (i.e., those with \(\widetilde{c}_i = 0\)); these can be written as
\begin{equation} \label{eq:activeSet}
\bm{U}\,\delta\bm{u}_k - \bm{W} = \bm{0}.
\end{equation}
The necessary conditions for optimality are derived from the Lagrangian
\begin{equation} \label{eq:lagrangian}
\begin{aligned}
    \mathcal{L}\bigl(\delta\bm{u}_k,\bm{\lambda}_k\bigr) &= \frac{1}{2}\delta\bm{u}_k^T\bm{Q}_{\bm{uu},k}\delta\bm{u}_k + \bm{Q}_{\bm{u},k}^T\delta\bm{u}_k \\
    &+ \bm{\lambda}_k^T\Bigl(\bm{U}\,\delta\bm{u}_k-\bm{W}\Bigr).
\end{aligned}
\end{equation}
Differentiating \eqref{eq:lagrangian} with respect to \(\delta\bm{u}_k\) and \(\bm{\lambda}_k\) yields
\begin{equation} \label{eq:langSystem}
\begin{bmatrix}
\bm{Q}_{\bm{uu},k} & \bm{U}\\
\bm{U}^T & \bm{0}
\end{bmatrix}
\begin{bmatrix}
\delta\bm{u}_k\\
\bm{\lambda}_k
\end{bmatrix}
=
\begin{bmatrix}
-\bm{Q}_{\bm{u},k}\\
\bm{W}
\end{bmatrix}.
\end{equation}
If we assume that the active constraints remain valid for any \(\bm{x}_k=\bar{\bm{x}}_k+\delta\bm{x}_k\), then
\begin{equation}
\bm{U}\,\delta\bm{u}_k-\bm{W}-\bm{X}\,\delta\bm{x}_k=\bm{0}.
\end{equation}
which leads to an affine control law
\begin{equation} \label{eq:controlLaw}
\delta\bm{u}_k=\bm{k}_k+\bm{K}_k\,\delta\bm{x}_k.
\end{equation}

\section{Model Predictive Control Framework}
To achieve real-time performance and safe trajectory planning under dynamically changing conditions, our approach integrates the DDP-based optimal control solution within an MPC framework. In this framework, the OCP augmented with the sigmoid safety boundaries decided by LLM is solved repeatedly over a receding horizon. At each control cycle, the LLM processes updated sensor and vehicle data to recompute the $\lambda$ values of the obstacles accordingly. These dynamic updates reshape the safety constraints, enabling the MPC to recalculate the vehicle's trajectories. Due to the computational efficiency of DDP, the MPC framework ensures real-time trajectory calculation. The closed-loop procedure maintains the safety requirements and enhances the overall driving efficiency by adapting to the complex, mixed-autonomy environments.

\begin{figure*}[ht]
    \centering
    \includegraphics[width=1\textwidth]{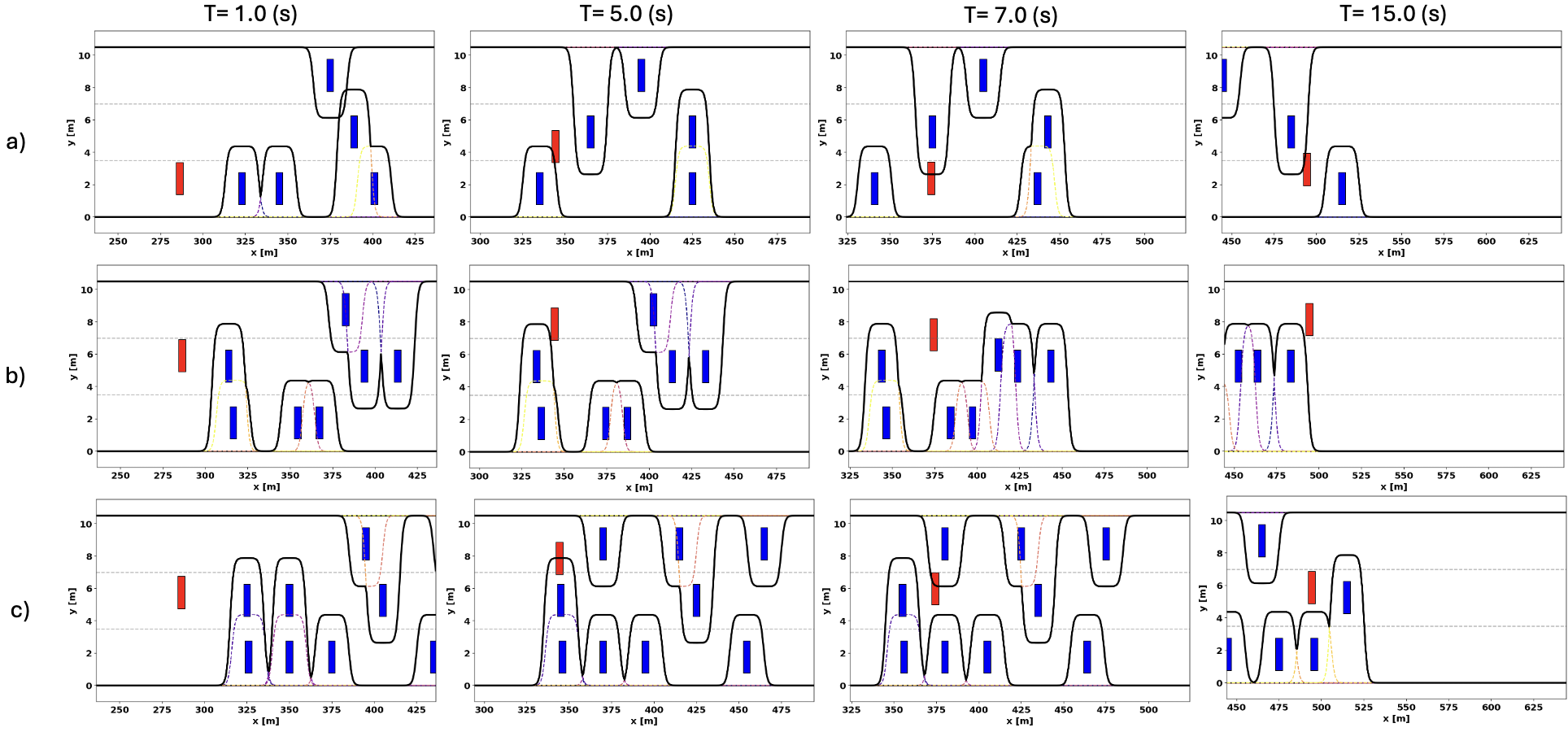}
    \caption{The simulation results of ego vehicle (red color) trajectories equal to 1.0 s, 5.0 s, 7.0 s, and 10.0 s with 5, 7, and 9 obstacle vehicles (blue color), obtained using \textit{CorrA}. The sigmoid-based boundaries are shown as black curves.}
    \label{fig:trajectory-time}
\end{figure*}

\section{Experiments}
 We conduct simulations with multiple lanes to validate our method. In the simulations, different traffic densities and vehicle initial positions are considered. The proposed approach is compared to a baseline MPC approach, which uses a combination of a feasible direction algorithm and a dynamic programming algorithm to solve the path planning problem (see \cite{typaldos2022optimization} for more details). For the LLMs, Deepseek-v3, deepseek-r, o3-mini and o1-mini are employed \cite{liu2024deepseek}. The temperature parameter of LLMs is set to 0 to ensure consistent results, where 0 ensures the outputs are more deterministic, while 1 allows for more creative responses.

\subsection{Metrics}
To evaluate the performance of CorrA, we adopt several metrics: (i) \textbf{Successful Rate}, which represents the percentage of scenarios in which the output of LLM did not lead the ego vehicle to collide with other obstacle vehicles; (ii) \textbf{Computation Time}, comparing the time needed for both DDP and the baseline approach to find the optimal solution; and (iii) \textbf{Travel Efficiency}, we measure how efficiently each control method performs by comparing their corresponding optimal trajectories, e.g., showing how effectively each method reaches and retains the vehicle's desired speed.

\subsection{Quantitative Evaluation}
To verify the consistency output of LLMs, we compare 2-lane and 3-lane scenarios using four different models, 20 runs individually. Deepseek-r and o3-mini are equipped with advanced reasoning ability by utilizing chain-of-thought, while deepseek-chat and o1-mini are zero-shot. Table \ref{tab: LLM} shows that a more advanced model is needed as the scenario becomes more complex from 2-lane to 3-lane. Specifically, for 2-lane scenarios, the successful rates of all models are over 90\%, while deep seek-chat and 01-mini are barely able to handle 3-lane scenarios. 
\begin{table}[h]
\caption{successful rate using different LLM models}
\label{tab: LLM}
\begin{center}
\begin{tabular}{l c c c c}
\hline
      & deepseek-r & deepseek-chat & o3-mini & o1-mini\\
\hline
2-lane & 100\% & 92\% & 100\% & 90\%\\
3-lane & 95\% & 5\% & 100\% & 5\%\\
\hline
\end{tabular}
\end{center}
\end{table}

The computational times required by both DDP and the baseline approach to converge to the optimal solution for different planning horizons are presented in Table \ref{tab: cpuTimes}. For shorter horizons, the performance gap between the two methods is negligible, with the baseline approach requiring 0.06 seconds and CorrA only needing 0.01 seconds. However, as the horizon increases, the difference rises significantly, and the computational advantage of CorrA becomes more profound. This trend arises by the fact that DDP's computational complexity grows linearly with the time horizon, showcasing the remarkable efficiency of DDP for real-time applications.

\begin{table}[ht]
\caption{CPU time comparison between CorrA and the baseline approach for different time horizons with $T=0.25$}
\label{tab: cpuTimes}
\begin{center}
\begin{tabular}{l c c c c}
\hline
    & 6 seconds & 8 seconds & 10 seconds & 12 seconds\\
\hline
Baseline & 0.06 & 0.1 & 0.15 & 0.21 \\
\textit{CorrA} & 0.01 & 0.01 & 0.01 & 0.02 \\
\hline
\end{tabular}
\end{center}
\end{table}

\begin{figure}[ht]
      \centering
      \includegraphics[width=1\linewidth]{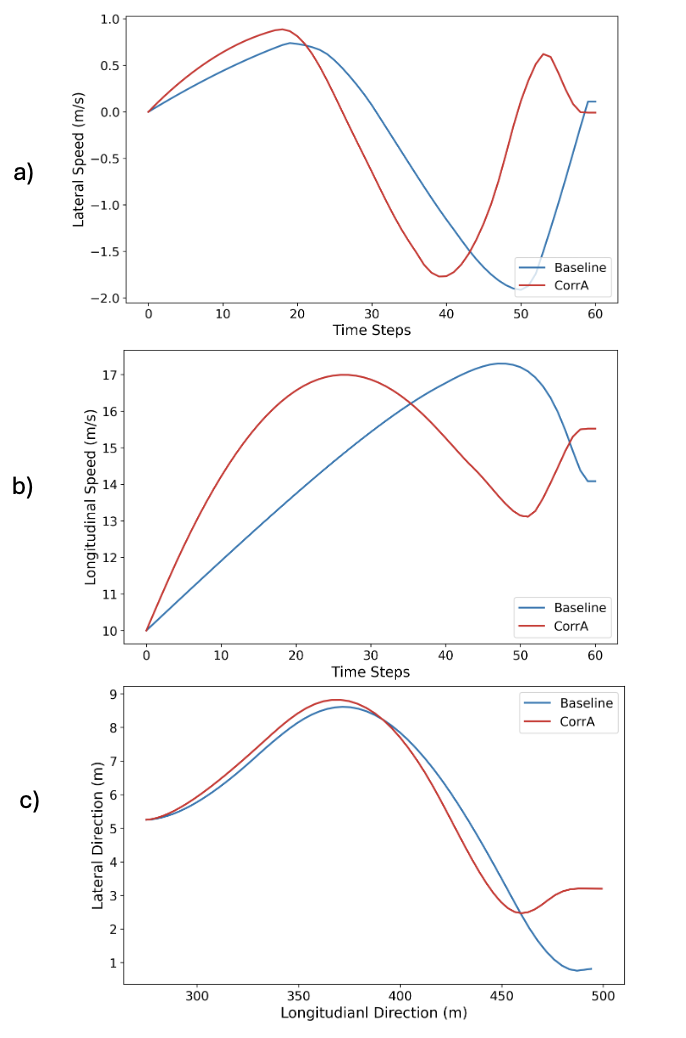}
      \caption{Optimal longitudinal and lateral speed and position trajectories of \textit{CorrA} versus the baseline approach.}
      \label{fig:speedVS}
   \end{figure}

\subsection{Qualitative Evaluation}
The dynamic evolution of safe corridors and vehicle trajectories generated by CorrA across multiple time steps and varying obstacle densities are illustrated in Fig. \ref{fig:trajectory-time}. The visualizations clearly demonstrate how the sigmoid-based boundaries adaptively form around obstacles, creating safe navigation corridors for the ego vehicle. Particularly, it can be seen how the boundaries dynamically reshape as the scenario changes, with the LLM continuously reassessing the optimal $\lambda$ values to maximize available space. This adaptive behavior is especially evident in complex scenarios with multiple obstacle vehicles in adjacent lanes, where CorrA successfully identifies and prioritizes the most efficient path while maintaining safety constraints, even in high-density traffic situations.

Figure \ref{fig:speedVS} shows the evolution of longitudinal and lateral speed and position trajectories over time for CorrA (red lines) compared to the baseline approach (blue lines). It can be seen that CorrA is able to maintain speeds closer to the desired while efficiently maneuvering through the surrounding traffic. In contrast, the baseline method achieves slower convergence to the desired speed and shows delayed maneuvering around the obstacles, as indicated in Fig. 5c. This comparison indicates CorrA's superior ability to adapt to the changing traffic environment.

\begin{figure}
    \centering
    \includegraphics[width=0.95\linewidth]{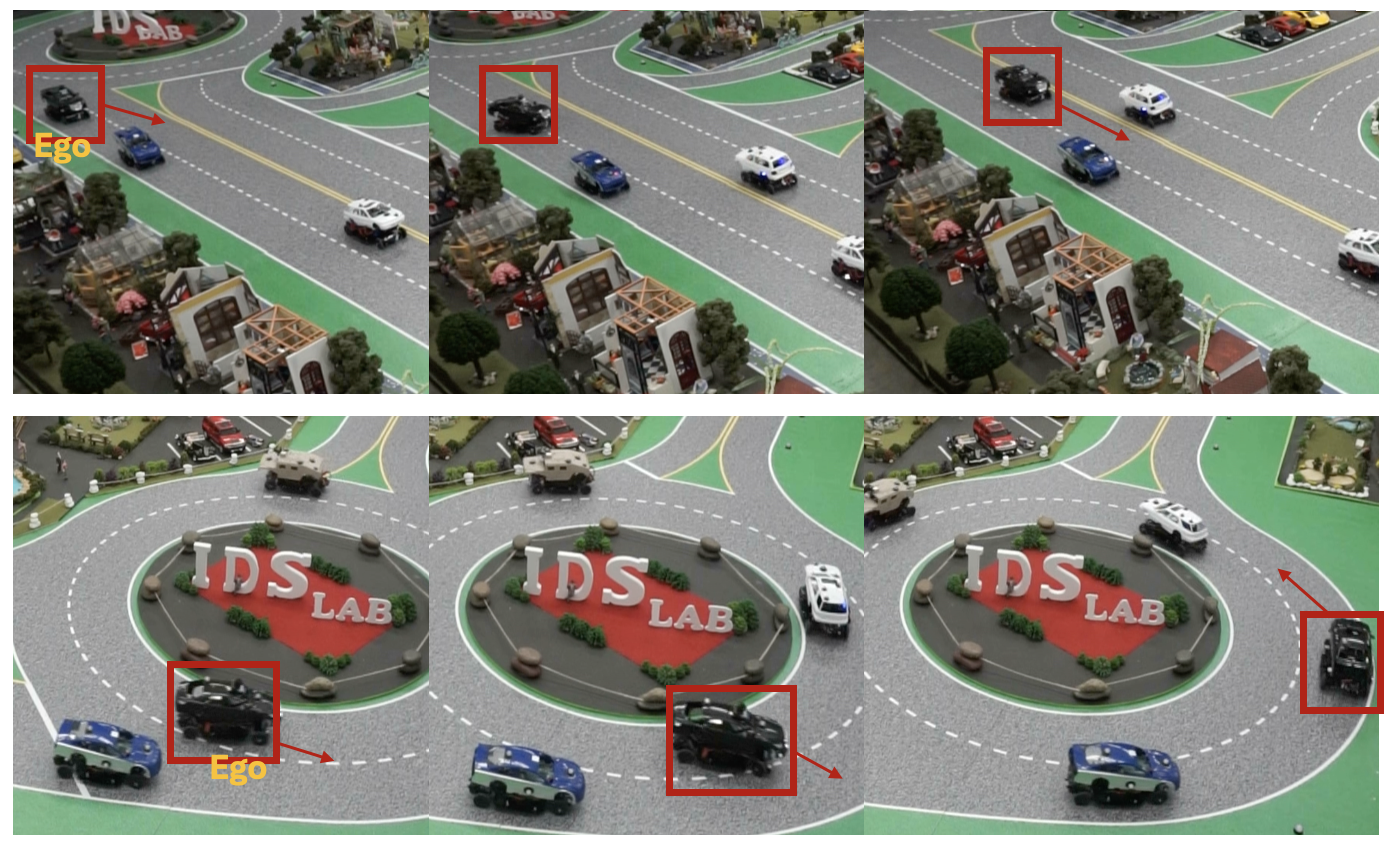}
    \caption{Experiment conducted in the IDS Lab's Scaled City. The black car marked by the red box is the ego car, which aims to maneuver through the surrounding vehicles. After the Vicon cameras capture information from all the cars, the ego car uses CorrA to compute its optimal trajectory. In this scenario, the ego car made two lane changes to overtake the downstream cars.}
    \label{fig:scale_city}
\end{figure}

\subsection{Practical Implementation and Experimental Validation}
We further validated CorrA through real-world implementation on robotic vehicles in the IDS Lab's Scaled City (IDS3C) facility, as Fig. \ref{fig:scale_city} shows. IDS3C provides a realistic 1:25 scale urban environment equipped with a high-precision Vicon motion capture system capable of millimeter-level tracking accuracy at 100Hz. The facility features a reconfigurable road network with various intersection types, traffic signals, and lane configurations that closely mimic real-world driving scenarios \cite{chalaki2022research}. Our experiments utilized autonomous robotic vehicles equipped with onboard computers running the CorrA framework. These physical tests support our simulation findings, with CorrA successfully handling dynamic obstacle avoidance. Notably, the LLM-based sigmoid boundary generation proved robust to the sensor noise and physical constraints inherent in real-world implementations, demonstrating successful navigation with zero collisions across 10 test runs of varying complexity. This experimental validation in a controlled physical environment provides strong evidence for CorrA's potential application in full-scale autonomous vehicles.

\section{Conclusions} \label{sec:conclusions}
In this paper, we introduced CorrA, a framework that combines the real-time optimal control of MPC with the reasoning capabilities of LLMs to tackle dynamic obstacle avoidance in autonomous driving. CorrA generates adaptive sigmoid-based safety boundaries around obstacles, effectively creating safe corridors that reduce the vehicle’s search space and prevent collisions. Simulations and scaled physical experiments validated CorrA’s effectiveness — the vehicle consistently maintained safe, efficient trajectories in complex, mixed-traffic scenarios, outperforming a baseline MPC in both speed of computation and travel efficiency. CorrA’s success in these diverse evaluations indicates strong potential for real-world deployment in full-scale AVs. 

Future research includes extending our work with the vision language model to formulate 3-D safety boundaries for more generalized application scenarios.



\bibliographystyle{IEEEtranS} 
\bibliography{IEEEabrv, IEEEbcpat, myIEEEbib, IDS_Publications}

\end{document}